\documentclass[runningheads]{llncs}
\usepackage[T1]{fontenc}
\usepackage{hyperref}
\hypersetup{
    colorlinks=true,
    linkcolor=blue,
    filecolor=magenta,      
    urlcolor=cyan
    }
\usepackage{graphicx}
\usepackage{algorithm}
\usepackage{algorithmicx}
\usepackage[noend]{algpseudocode}
\usepackage{booktabs}
\usepackage{amsmath}
\usepackage{amsfonts}
\usepackage{amssymb}
\usepackage{multirow}
\usepackage{xcolor}
\usepackage{pifont}
\usepackage{listings}
\usepackage{varwidth}
\usepackage{float}

\definecolor{gray}{rgb}{0.4,0.4,0.4}
\definecolor{darkblue}{rgb}{0.0,0.0,0.6}
\definecolor{cyan}{rgb}{0.0,0.6,0.6}
\definecolor{maroon}{rgb}{0.5,0,0}
\definecolor{darkgreen}{rgb}{0,0.7,0}
\definecolor{darkred}{rgb}{0.7,0,0}
\lstdefinestyle{customxml}{
  language=XML,
  basicstyle=\scriptsize\ttfamily,
  columns=fullflexible,
  showstringspaces=false,
  commentstyle=\color{gray}\upshape,
  morestring=[s]{"}{"},
  morecomment=[s]{?}{?},
  morecomment=[s]{!--}{--},
  commentstyle=\color{darkgreen},
  moredelim=[s][\color{black}]{>}{<},
  moredelim=[s][\color{orange}]{\ }{=},
  stringstyle=\color{cyan},
  identifierstyle=\color{cyan}
}

\lstdefinestyle{pythonstyle}{
    language=Python,
    basicstyle=\ttfamily\small,
    keywordstyle=\color{blue}\bfseries,
    commentstyle=\color{green!40!black},
    stringstyle=\color{orange},
    showstringspaces=false,
    tabsize=4,
    backgroundcolor=\color{gray!3},
    captionpos=b,
    breakatwhitespace=false,
    breaklines=true,
    captionpos=b,
    keepspaces=true,
    showspaces=false,
    showstringspaces=false,
    showtabs=false,
    xleftmargin=15pt,
    xrightmargin=15pt,
    numbers=none, 
    frame=none,   
}

\newcommand{\cmark}{\textcolor{darkgreen}{\ding{51}}}
\newcommand{\xmark}{\textcolor{darkred}{\ding{55}}}
\newcommand{\dashed}{--} 
\begin{document}
\title{PyGraft: Configurable Generation of Synthetic Schemas and Knowledge Graphs at Your Fingertips}

\titlerunning{PyGraft: Configurable Generation of Synthetic Schemas and KGs}

\author{
Nicolas Hubert\inst{1,2}\orcidID{0000-0002-4682-422X}\and
Pierre Monnin\inst{3}\orcidID{0000-0002-2017-8426}\and
Mathieu d'Aquin\inst{2}\orcidID{0000-0001-7276-4702}
\and
Davy Monticolo\inst{1}\orcidID{0000-0002-4244-684X}
\and
Armelle Brun\inst{2}\orcidID{0000-0002-9876-6906}
}
\authorrunning{N. Hubert et al.}

\institute{Université de Lorraine, ERPI, Nancy, France \and
Université de Lorraine, CNRS, LORIA, Nancy, France \and
Université Côte d’Azur, Inria, CNRS, I3S, Sophia-Antipolis, France
\email{\{nicolas.hubert,mathieu.daquin,armelle.brun,davy.monticolo\}@univ-lorraine.fr}
\email{pierre.monnin@inria.fr}}
\authorrunning{N. Hubert et al.}
%
%
\maketitle              
\begin{abstract}
Knowledge graphs (KGs) have emerged as a prominent data representation and management paradigm. Being usually underpinned by a schema (\textit{e.g.,} an ontology), KGs capture not only factual information but also contextual knowledge. In some tasks, a few KGs established themselves as standard benchmarks. However, recent works outline that relying on a limited collection of datasets is not sufficient to assess the generalization capability of an approach. In some data-sensitive fields such as education or medicine, access to public datasets is even more limited.
To remedy the aforementioned issues, we release PyGraft, a Python-based tool that generates highly customized, domain-agnostic schemas and KGs. The synthesized schemas encompass various \texttt{RDFS} and \texttt{OWL} constructs, while the synthesized KGs emulate the characteristics and scale of real-world KGs.
Logical consistency of the generated resources is ultimately ensured by running a description logic (DL) reasoner. By providing a way of generating both a schema and KG in a single pipeline, PyGraft's aim is to empower the generation of a more diverse array of KGs for benchmarking novel approaches in areas such as graph-based machine learning (ML), or more generally KG processing. In graph-based ML in particular, this should foster a more holistic evaluation of model performance and generalization capability, thereby going beyond the limited collection of available benchmarks. PyGraft is available at: \url{https://github.com/nicolas-hbt/pygraft}.

\end{abstract}

\noindent \textbf{Resource type:} Software

\noindent \textbf{License:} MIT License

\noindent \textbf{DOI:} \url{https://doi.org/10.5281/zenodo.10243209}

\noindent \textbf{URL:} \url{https://github.com/nicolas-hbt/pygraft}

\section{Introduction}

Knowledge graphs (KGs) have been increasingly used as a graph-structure representation of data. More specifically, a KG is a collection of triples $(s,p,o)$ where $s$ (subject) and $o$ (object) are two entities of the graph, and $p$ is a predicate that qualifies the nature of the relation holding between them~\cite{kgbook}.
KGs are usually underpinned by a schema (\textit{e.g.,} an ontology) which defines the main concepts and relations of a domain of interest, as well as the rules under which these concepts and relations are allowed to interact~\cite{gruber1995}.

KGs are being used in a wide array of tasks, in many of which a limited collection of KGs established themselves as standard benchmarks for evaluating model performance. However, there are some concerns around the sole usage of these specific mainstream KGs for assessing the generalization capability of newly introduced models. For example, for the particular task of node classification, it has been demonstrated that mainstream datasets such as CiteSeer, Cora, and PubMed feature similar statistical characteristics, especially homophily~\cite{palowitch2022}. Consequently, new models are assessed with respect to a collection of statistically similar datasets. Therefore, their performance improvement does not always hold beyond the standard benchmark datasets~\cite{palowitch2022}.
Similarly, it has been shown that many link prediction train sets suffer from extremely skewed distributions in both the degrees of entities and the occurrence of a subset of predicates~\cite{rossi2020relations}. In addition, some of the established link prediction datasets are plagued with data biases~\cite{rossi2021} and include many occurrences of inference patterns~\cite{dettmers2018,jin2023,liu2023} that predictive models are able to incorporate, which may cause overly optimistic evaluation performance~\cite{dettmers2018}. 
For example, the mainstream FB15k and WN18 datasets are substantially affected by these biases.
Therefore, more diverse datasets are needed~\cite{rossi2021}.
In this situation, it is of utmost importance to provide a way for researchers to generate synthetic yet realistic datasets of different shapes and characteristics, so that new models can be evaluated in a wide range of data settings.

Worse than relying on a limited number of KGs is the lack of publicly available KGs in some application fields. Conducting research in domains such as education, law enforcement, or medicine, is particularly difficult. On the grounds of data privacy, collecting and sharing real-world knowledge may not be possible. As such, domain-oriented KGs are barely accessible in these areas. However, engineers, practitioners, and researchers usually have precise ideas of the characteristics of their problem of interest. In this context, it would be beneficial to generate a synthetic KG that emulates the characteristics of a real KG~\cite{feng2021}. 

The aforementioned issues led to several attempts at building synthetic generators of schemas and KGs, even though in most cases these two aspects have been considered separately.
Stochastic-based generators have been proposed to output domain-agnostic KGs~\cite{barabasi2002,erdos1959,trilliong2017}. However good these methods are at generating large graphs quickly, their data generation process does not allow to take an underlying schema into account~\cite{feng2021}. Therefore, the resulting KGs are not guaranteed to accurately mimic the characteristics of real-world KGs in a desired application field.
In contrast, schema-driven generators are able to synthesize KGs that resemble real-world data. However, most works focused on generating synthetic KGs on the basis of an already existing schema~\cite{melo2017}. Synthesizing both a schema and a KG underpinned by it is a more challenging task that has been considered but with only limited success so far~\cite{melo2017}. 

In this work, we aim at addressing this issue.
In particular, we present PyGraft, a Python-based tool to generate highly customized, domain-agnostic schemas and KGs.
The contributions of our work are the following:
\begin{itemize}
    \item To the best of our knowledge, PyGraft is the first generator dedicated to the generation of both schemas and KGs in a unique pipeline, while being highly tunable based on a broad array of user-specified parameters. Notably, the generated resources are domain-agnostic, which makes them usable for benchmarking purposes regardless of the application field.
    \item The generated schemas and KGs are built with an extended set of \texttt{RDFS} and \texttt{OWL} constructs and their logical consistency is ensured by the use of a DL reasoner, which allows for both fine-grained description of resources and strict compliance with common Semantic Web standards.
    \item We publicly release our code with a documentation and accompanying examples for ease of use.
\end{itemize}

The remainder of the paper is structured as follows. 
Related work is presented in Section~\ref{related-work}. In Section~\ref{pygraft-description}, PyGraft is detailed and accompanied with necessary background knowledge and comparisons with other available generators. In Section~\ref{pygraft-use-case}, a thorough performance analysis of PyGraft is presented. A discussion on current potential use cases, limitations, and future work can be found in Section~\ref{discussion}.
Lastly, Section~\ref{conclusion} sums up the key insights presented in this paper.

\section{Related Work}\label{related-work}
The generation principle governing synthetic graph generators leads to differentiate between \emph{stochastic-based}, \emph{deep generative}, and \emph{semantic-driven} generators. Along the description of these families of generators, Table~\ref{tab:libraries} provides details on current and open-source implementations. These tools are also compared with PyGraft regarding several criteria.

\textbf{Stochastic-based} generators are usually characterized by their ability to output large graphs in a short amount of time. Early works are represented by the famous Erdős–Rényi model~\cite{erdos1959}. 
The Erdős–Rényi model generates graphs by independently assigning edges between pairs of nodes with a fixed probability. The Barabási-Albert model~\cite{barabasi2002} exhibits scale-free degree distributions and is based on the principle of preferential attachment, where new nodes are more likely to attach to nodes with higher degrees. The R-MAT model~\cite{rmat2004}
generates large-scale power-law graphs with properties like power-law degree distributions and community structures. More recently, TrillionG~\cite{trilliong2017} -- presented as an extension of R-MAT -- represents nodes and edges as vectors in a high-dimensional space. TrillionG allows users to generate large graphs up to trillions of edges while exhibiting lower space and time complexities than previously proposed generators.
In Table~\ref{tab:libraries}, stochastic-based generators are represented by igraph\footnote{\url{https://github.com/igraph/python-igraph/}}, NetworkX\footnote{\url{https://github.com/networkx/networkx/}}, and Snap\footnote{\url{https://github.com/snap-stanford/snap-python/}}. Although they are not specifically designed for graph generation, they provide off-the-shelf implementations for generating random graphs, such as the Erdős–Rényi and Barabási–Albert models. They are domain-agnostic and scalable, but do not take into account any semantics.

\textbf{Deep generative} graph generators are trained on existing graph datasets and learn to capture the underlying patterns of the input graphs. Deep generative graph models are typically based on generative adversarial networks (GANs) and graph neural networks (GNNs), recurrent neural networks (RNNs), or variational autoencoders (VAEs).
GraphGAN~\cite{graphgan2018} leverages the GAN structure, in which the generative model receives a vertex and aims at fitting its true connectivity distribution over all other vertices -- thereby producing fake samples for the discriminative model to differentiate from ground-truth samples. GraphRNN~\cite{graphrnn2018} is a deep autoregressive model that trains on a collection of graphs. It
can be viewed as a hierarchical model adding nodes and edges in a sequential manner.
A representant of the VAE family of generators is NeVAE~\cite{nevae2020}, which is specifically designed for molecular graphs. NeVAE features a decoder which is able to guarantee a set of valid properties in the generated molecules.
In particular, MolGAN~\cite{molgan2018} and NeVAE~\cite{nevae2020} are bound to molecular graph generation. Therefore, they are not domain-agnostic (Table~\ref{tab:libraries}). The other generators of this family are, but they do not take a schema as input when generating random graphs, \textit{i.e.,} they are not schema-driven (Table~\ref{tab:libraries}).

\textbf{Semantic-driven} generators, in contrast, incorporate schema-based constraints or external knowledge to generate graphs that exhibit specific characteristics or follow certain patterns relevant to the given field of application.
In~\cite{guo2005}, the Lehigh University Benchmark (LUBM) and the Univ-Bench Artificial data generator (UBA) are presented. The former is an ontology modelling the university domain while the latter aims at generating synthetic graphs based on the LUBM schema as well as user-defined queries and restrictions. Similarly, the Linked Data Benchmark Council (LDBC)~\cite{angles2014} released the Social Network Benchmark (SNB), which includes a graph generator for synthesizing social network data based on realistic distributions. gMark~\cite{bagan2017} has subsequently been presented as the first generator that satisfies the criteria of being domain-independent, scalable, schema-driven, and highly configurable, all at the same time. However, it still requires already existing schemas as input.
In~\cite{melo2017}, Melo and Paulheim focus on the synthesis of KGs for the purpose of benchmarking link prediction and type prediction tasks. The authors claim that there is a need for more diverse benchmark datasets for link prediction, with the possibility of having control over their characteristics (\textit{e.g.,} the number of entities, relation assertions, number of types, etc.). Therefore, Melo and Paulheim propose a synthesizing approach which replicates input real-world graphs while allowing for controlled variations in graph characteristics. Notably, they highlight the fact that most works, including theirs, focus on synthesizing KGs based on an existing schema, which leads them to formulate the desiderata of generating both a schema and a KG from scratch as a promising venue for future work -- which PyGraft actually does. Subsequently, Feng \textit{et al.}~\cite{feng2021} proposed a schema-driven graph generator based on the concept of Extended Graph Differential Dependencies ($GDD^{x}$). However, their approach cannot generate domain-agnostic schemas and thus requires an existing schema as input. The DLCC benchmark proposed in~\cite{portisch2022} features a synthetic KG generator based on user-specified graph and schema properties. Beyond asking for a given number of nodes, relations and degree distribution in the resulting KG, it allows for specifying a few \texttt{RDFS} constructs for the generation of the underpinning schema.

None of the aforementioned semantic-driven generators perform a logical consistency check of the generated graphs (see Table~\ref{tab:libraries}). Additionally, they can only produce final KGs based on an input schema. 
Some of them are also domain-specific.
DLCC~\cite{portisch2022} is the closest work to ours, and, to the best of our knowledge, this is the first and only work that allows to generate both a schema and a KG while being domain-agnostic. However, it only features 3 schema constructs, namely \texttt{rdfs:domain}, \texttt{rdfs:range}, and \texttt{rdfs:subClassOf}. These sole three constructs do not pose any constraints on triple generation (hence the consistency checking is not needed) and do not fully feature all Semantic Web possibilities that could exist in KGs. This is also why the resources generated with the DLCC generator do not undergo any logical consistency checks (see Table~\ref{tab:libraries}).

In the present work, we aim at going a step further and taking numerous \texttt{RDFS} \emph{and} \texttt{OWL} constructs into account, as PyGraft features a broad range of schema constructs (see Tables~\ref{tab:parameters} and \ref{tab:relation-properties}). Considered together, these constructs lead to many potential sources of inconsistencies that need to be carefully avoided with specific procedures and a final consistency check using a DL reasoner. However, this more challenging setting allows PyGraft to generate KGs that feature a broader range of Semantic Web possibilities.

\begin{table}[]
\caption{Feature comparison of graph generation tools. Dashed line is used when a feature is not applicable due to the characteristics of the described generation tool. Domain-agnostic denotes whether a given tool is able to potentially operate with schemas of different application fields.}
\label{tab:libraries}
\centering
\setlength{\tabcolsep}{0.15cm}
\resizebox{1.0\textwidth}{!}{
\begin{tabular}{cccccccc}
\toprule 
Tool & Domain-agnostic & Schema-driven & Schema generation & Schema properties & Graph properties & Scalable & Consistency check \\
\midrule
igraph\footnote{\url{https://github.com/igraph/python-igraph/}} & \cmark & \xmark & \xmark & \dashed & \cmark & \cmark & \dashed \\
\midrule
NetworkX\footnote{\url{https://github.com/networkx/networkx/}} & \cmark & \xmark & \xmark & \dashed & \cmark & \cmark & \dashed \\
\midrule
Snap\footnote{\url{https://github.com/snap-stanford/snap-python/}} & \cmark & \xmark & \xmark & \dashed & \cmark & \cmark & \dashed \\
\midrule
GraphGen~\cite{graphgen2020} & \cmark & \xmark & \xmark & \dashed & \cmark & \cmark & \dashed \\
\midrule
GraphRNN~\cite{graphrnn2018} & \cmark & \xmark & \xmark & \dashed & \cmark & \cmark & \dashed \\
\midrule
GraphVAE~\cite{graphvae2018} & \cmark & \xmark & \xmark & \dashed & \cmark & \xmark & \dashed \\
\midrule
GraphWorld~\cite{palowitch2022} & \cmark & \xmark & \xmark & \dashed & \cmark & \cmark & \dashed \\
\midrule
MolGAN~\cite{molgan2018} & \xmark & \xmark & \xmark & \dashed & \cmark & \xmark & \dashed \\
\midrule
NeVAE~\cite{nevae2020} & \xmark & \xmark & \xmark & \dashed & \cmark & \cmark & \dashed \\
\midrule
UBA-LUBM~\cite{guo2005} & \xmark & \cmark & \xmark & \dashed & \cmark & \cmark & \xmark \\
\midrule
SNB~\cite{angles2014} & \xmark & \cmark & \xmark & \dashed & \cmark & \cmark & \xmark \\
\midrule
gMark~\cite{bagan2017} & \cmark & \cmark & \xmark & \dashed & \cmark & \cmark & \xmark \\
\midrule
Melo \textit{et al.}~\cite{melo2017} & \cmark & \cmark & \xmark & \dashed & \cmark & \cmark & \xmark \\
\midrule
$GDD^{x}$~\cite{feng2021} & \cmark & \cmark & \xmark & \dashed & \cmark & \cmark & \xmark \\
\midrule
DLCC~\cite{portisch2022} & \cmark & \cmark & \cmark & $3$ & \cmark & \cmark & \xmark \\
\midrule
PyGraph (ours) & \cmark & \cmark & \cmark & $13$ & \cmark & \cmark & \cmark \\
\bottomrule
\end{tabular}}
\end{table}

\section{PyGraft Description}\label{pygraft-description}
This section starts by formally introducing the notions of schema and knowledge graph. The schematic overview of PyGraft is presented in Section~\ref{overview}. PyGraft schema and KG generation pipelines are presented in Sections~\ref{schema-generation} and \ref{kg-generation}, respectively.

\subsection{Preliminaries}
On the one hand, a schema -- \textit{e.g.,} an ontology -- refers to a explicit specification of a conceptualization that includes concepts, properties, and restrictions within a particular domain of knowledge~\cite{gruber1995}. It helps ensure consistency, clarity, and interoperability when representing and sharing knowledge.
In our work, we consider schemas to be represented as a collection of concepts $\mathcal{C}$, properties $\mathcal{P}$, and axioms $\mathcal{A}$, \textit{i.e.,} $\mathcal{S} = \{ \mathcal{C}, \mathcal{P}, \mathcal{A}\}$.
Schemas are typically represented using formal languages such as \texttt{RDFS}\footnote{\url{https://www.w3.org/RDFS/}} (Resource Description Framework) and \texttt{OWL}\footnote{\url{https://www.w3.org/OWL/}} (Web Ontology Language).

Regarding KGs, distinct definitions co-exist~\cite{bonatti,ehrlinger}. In this work, we stick to the inclusive definition of Hogan \textit{et al.}~\cite{kgbook}, \textit{i.e.,} we consider a KG to be a graph where nodes represent entities and edges represent relations between these entities.
The link between schemas and KGs lies in the fact that schemas are often used to define the structure and semantics of a KG. In other words, a schema defines the vocabulary and rules that govern entities and relationships in a KG.
In this view, a KG is a data graph that can be potentially enhanced with a schema~\cite{kgbook}.

\subsection{Overview}\label{overview}
\begin{figure}[]
  \centering
  \includegraphics[scale=0.6]{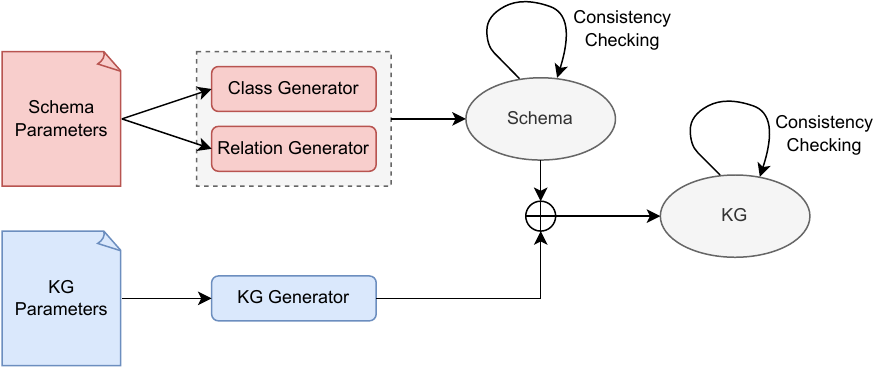}
  \caption{PyGraft general overview.}
  \label{fig:pygraft-overview}
\end{figure}
From a high-level perspective, the entire PyGraft generation pipeline is depicted in Fig.~\ref{fig:pygraft-overview}. In particular, Class and Relation Generators are firstly initialized with user-specified parameters, and are then used for building the schema incrementally. The logical consistency of the schema is subsequently checked using HermiT reasoner~\cite{hermit} through the \texttt{owlready2}\footnote{\url{https://github.com/pwin/owlready2/}} Python library. If the user is also interested in generating a KG based on this schema, the KG Generator is initialized with KG-related parameters, and takes the previously generated schema as input in order to sequentially build the KG. Ultimately, the logical consistency of the resulting KG is assessed with HermiT.
More details on the schema-level generation are provided in Section~\ref{schema-generation}, while Section~\ref{kg-generation} describes the KG generation procedure.

\subsection{Schema Generation}\label{schema-generation}
\begin{table}[]
\caption{User-defined parameters for schema and KG generations.}\label{tab:parameters}
\centering
\footnotesize
\setlength{\tabcolsep}{0.25cm}
\resizebox{1.0\textwidth}{!}{
\begin{tabular}{lll}
\toprule 
& Parameter & Description\\
\midrule
\multirow{5}{*}{{Classes}} & \texttt{num\_classes} & Number of classes \\
& \texttt{max\_depth} & Depth of the class hierarchy \\
& \texttt{avg\_depth} & Average class depth \\
& \texttt{inheritance\_ratio} & Proportion of \texttt{rdfs:subClassOf} \\
& \texttt{avg\_disjointness} & Proportion of \texttt{owl:DisjointWith} \\
\midrule
\multirow{12}{*}{{Relations}} & \texttt{num\_relations} & Number of relations \\
& \texttt{prop\_profiled\_relations} & Proportion of \texttt{rdfs:domain} and \texttt{rdfs:range} \\
& \texttt{relation\_specificity} & Average depth of \texttt{rdfs:domain} and \texttt{rdfs:range} \\
& \texttt{prop\_asymmetric} & Proportion of \texttt{owl:AsymmetricProperty} \\
& \texttt{prop\_symmetric} & Proportion of \texttt{owl:SymmetricProperty} \\
& \texttt{prop\_irreflexive} & Proportion of \texttt{owl:IrreflexiveProperty} \\
& \texttt{prop\_reflexive} & Proportion of \texttt{owl:ReflexiveProperty} \\
& \texttt{prop\_transitive} & Proportion of \texttt{owl:TransitiveProperty} \\
& \texttt{prop\_functional} & Proportion of \texttt{owl:FunctionalProperty} \\
& \texttt{prop\_inversefunctional} & Proportion of \texttt{owl:InverseFunctionalProperty} \\
& \texttt{prop\_inverseof} & Proportion of \texttt{owl:inverseOf} \\
& \texttt{prop\_subproperties} & Proportion of \texttt{rdfs:subPropertyOf} \\
\midrule
\multirow{7}{*}{{Individuals}} & \texttt{num\_entities} & Number of entities \\
& \texttt{num\_triples} & Number of triples \\
& \texttt{relation\_balance} & Relation distribution across triples \\
& \texttt{prop\_untyped} & Proportion of untyped entities\\
& \texttt{avg\_depth\_specific} & Average depth of most specific class \\
& \texttt{multityping} & Whether entities are multi-typed \\
& \texttt{avg\_multityping} & Average number of most-specific classes per entity\\
\bottomrule
\end{tabular}
}
\end{table}

The schema generation follows a well-defined series of steps. In particular, the Class Generator in Fig.~\ref{fig:pygraft-overview} is initialized first, and generates the class hierarchy and disjointness axioms as detailed in Algorithm~\ref{alg:class-generation}. Then, the Relation Generator is initialized and handles the relation generation following the procedure of Algorithm~\ref{alg:relation-generation}. In the following, each algorithm is described step by step.

\textbf{Class Generation.} First, classes are generated based on the user-specified number of classes \texttt{num\_classes} (lines 1-2). Then, the user-specified \texttt{max\_depth} parameter is satisfied by taking one class after the other and creating child-parent connections through the \texttt{rdfs:subClassOf} assertion (lines 3-7). At this point, the class hierarchy is a purely vertical tree where each node (\textit{i.e.,} class) has exactly one child, except the leaf node. The class hierarchy is then further filled by taking each remaining class sequentially, and connecting it with other classes so that \texttt{avg\_depth} and \texttt{inheritance\_ratio} are satisfied (lines 8-13). It is worth mentioning that with some probability $\alpha$ chosen to be moderately low, a freshly picked class can be placed randomly (lines 10-11). This does not necessarily go in the direction of \texttt{avg\_depth} and \texttt{inheritance\_ratio} target parameters, but it allows adding stochasticity and realism in the characteristics of the generated class hierarchy. Besides, when a low number of classes are still to be connected, this randomness is deactivated so that each subsequent class connection is in line with \texttt{avg\_depth} and \texttt{inheritance\_ratio} fulfilment. Finally, class disjointnesses are added to the schema by picking two classes A and B, ensuring that none of them is a transitive parent or child of the other, and extending class disjointness to their respective children, if any (lines 14-16).

\begin{figure}[t]
  \centering
  \includegraphics[scale=0.595]{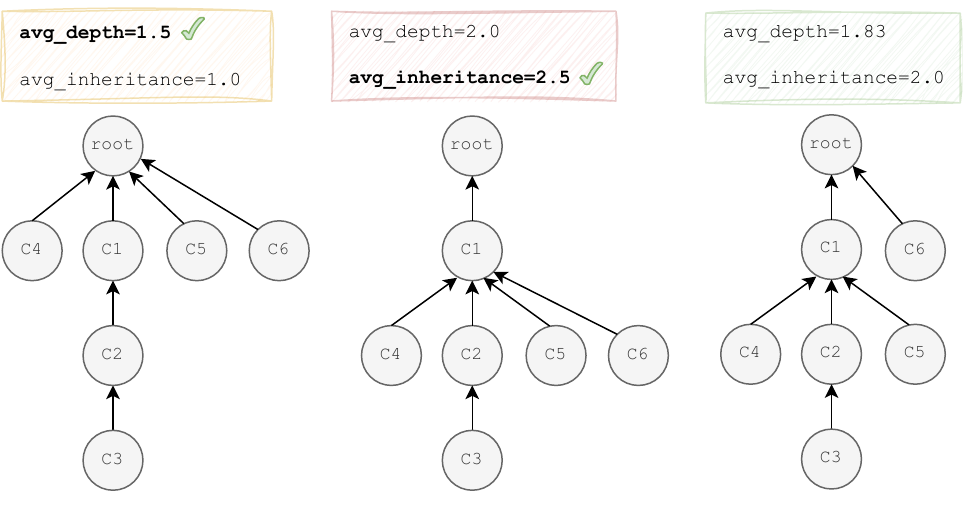}
  \caption{Potential class hierarchies for the constraints: $\texttt{num\_classes} = 6$, $\texttt{max\_depth} = 3$, $\texttt{avg\_depth} = 1.5$, and $\texttt{inheritance\_ratio} = 2.5$. Left and middle class hierarchies are built with parameter priority. The right class hierarchy is built with a best-effort strategy, without specific parameter privilege.}
  \label{fig:class-trees}
\end{figure}

When generating classes, anomalies may occur. For instance, choosing values such that $\texttt{avg\_depth} > \texttt{max\_depth}$ triggers an error. In a few other situations, the schema is generated but the user requirements might not be completely fulfilled. This can happen because of competing parameter values. For example, the constraints $\texttt{num\_classes} = 6$ (\texttt{root} excluded), $\texttt{max\_depth} = 3$, $\texttt{avg\_depth} = 1.5$, and $\texttt{inheritance\_ratio} = 2.5$ cannot be simultaneously satisfied (Fig.~\ref{fig:class-trees}). In this situation, the schema is generated with a best-effort strategy, seeking to build a schema with statistics as close as possible to the user requirements.

\textbf{Relation Generation.} Before presenting the procedure for generating relations and their properties, it is worth mentioning that PyGraft allows relations to be described by multiple \texttt{OWL} and \texttt{RDFS} constructs. This leads to more realistic schemas and KGs, at the expense of higher risk of inconsistency: some property combinations are not logically consistent, \textit{e.g.,} a relation cannot be simultaneously qualified by \texttt{owl:ReflexiveProperty} and \texttt{owl:IrreflexiveProperty}. Based on the relation properties available in PyGraft (Table~\ref{tab:relation-properties}), all combinations were extracted and for each combination, a new graph was serialized using \texttt{rdflib}\footnote{\url{https://github.com/RDFLib/rdflib/}}. This graph contains a unique relation which is qualified by the given property combinations\footnote{An instance triple should also be added. This is because some property combinations such as \texttt{owl:SymmetricProperty} and \texttt{owl:AsymmetricProperty} are not flagged as logically inconsistent \textit{per se} in \texttt{OWL}. However, a relation qualified by these two properties is not allowed to connect any instances.}. Based on the simplified schema, the HermiT reasoner performs consistency checking. Finally, a dictionary stores all possible property combinations. Knowing valid and invalid property combinations is necessary for guiding relation property assignment and minimize logical inconsistency likelihood before actually running the reasoner.

In Algorithm~\ref{alg:relation-generation}, this dictionary of valid property combinations is loaded (line 3) just after initializing the number of relations specified by the user (lines 1-2). Then, each relation in the schema is qualified with properties based on a pre-defined order (lines 4-6). For example, \texttt{owl:ReflexiveProperty} and \texttt{owl:IrreflexiveProperty} are assigned first, then \texttt{owl:SymmetricProperty}, etc. These properties are named \emph{attributes} as they characterize a relation \textit{per se}. Next, the \emph{relationship property} \texttt{owl:InverseOf} is assigned to relations (line 7), which poses constraints on relation domain and range assignments (line 8), which themselves pose constraints on relation pairings through the \texttt{rdfs:subPropertyOf} assertion (line 9), \textit{e.g.,} domain and range of subproperties should not be disjoint with domain and range of superproperties.

In Table~\ref{tab:relation-properties} are reported the relation properties that the current PyGraft version handles, along with their logical definition and accompanying examples.

\begin{table}[h]
\caption{Relation properties covered by PyGraft.}
\label{tab:relation-properties}
\centering
\setlength{\tabcolsep}{0.15cm}
\resizebox{1.0\textwidth}{!}{
\begin{tabular}{lll}
\toprule 
Property & Definition & Example\\
\midrule
\texttt{owl:AsymmetricProperty} & $\forall x \forall y: p(x, y) \Rightarrow \neg p(y, x)$ & \texttt{isParentOf} \\
\texttt{owl:SymmetricProperty} & $\forall x \forall y: p(x, y) \Rightarrow p(y, x)$ & \texttt{hasSibling} \\
\texttt{owl:ReflexiveProperty} & $\forall x: p(x, x)$ & \texttt{hasSameColorAs} \\
\texttt{owl:IrreflexiveProperty} & $\forall x: \neg p(x, x)$ & \texttt{isYoungerThan} \\
\texttt{owl:TransitiveProperty} & $\forall x \forall y \forall z: p(x, y) \land p(y, z) \Rightarrow p(x, z)$ & \texttt{isCheaperThan} \\
\texttt{owl:FunctionalProperty} & $\forall x \forall y \forall z: (p(x, y) \land p(x, z)) \Rightarrow y = z
$ & \texttt{hasISBN} \\
\texttt{owl:InverseFunctionalProperty} & $\forall x \forall y \forall z: (p(x, y) \land p(z, y)) \Rightarrow x = z
$ & \texttt{isEmailAddressOf} \\
\texttt{owl:InverseOf} & $\forall x \forall y: p(x, y) \Longleftrightarrow q(y, x)
$ & $\texttt{owns} \Longleftrightarrow \texttt{isOwnedBy}$ \\
\texttt{rdfs:subPropertyOf} & $\forall x \forall y: p(x, y) \Rightarrow q(x, y)
$ & $\texttt{hasMother} \Rightarrow \texttt{hasParent}$\\
\bottomrule
\end{tabular}
}
\end{table}

\subsection{Knowledge Graph Generation}\label{kg-generation}
In light of PyGraft overview (Fig.~\ref{fig:pygraft-overview}), this section explores how the KG generator is initialized with user parameters and used in conjunction with any generated schema to build the final KG. As for schema, we provide a step-by-step insight into Algorithm~\ref{alg:kg-generation}.
Entities and the KG are generated (lines 1-3) and a schema dedicated to the KG generation is loaded (line 4). Based on \texttt{prop\_untyped}, entities are assigned a class whose depth in the class hierarchy should be in a close range around \texttt{avg\_depth\_specific} (line 5). Based on \texttt{avg\_multityping}, several of them are assigned other classes of the same depth, provided that they are not disjoint with any of the specific classes characterizing a given entity (lines 6-7). Then, triples are generated (lines 8-9) in a sequential manner. Namely, unobserved entities in $U$ have sampling priority, to ensure that the required number of entities in the resulting KG is met as fully as possible. Ultimately, checking procedures are performed (line 10) before the KG undergoes logical consistency checking using a semantic reasoner (line 11). If the generated KG is inconsistent, a message warns the user but the KG is stored nevertheless. The user can then choose to restart the generation procedure.
It is worth mentioning that the checking procedures (line 10) follow the relevant \texttt{OWL 2 RL/RDF} rules\footnote{\url{https://www.w3.org/TR/owl2-profiles/\#ref-owl-2-rdf-semantics/}}. These rules -- in the form of first-order implications -- are implemented in PyGraft to ensure consistency pre-checking before the HermiT reasoner is deployed. In addition, the OntOlogy Pitfall Scanner!\footnote{\url{https://oops.linkeddata.es/}} was also used for identifying several checking procedures and for ensuring compliance with common Semantic Web standards.

\begin{figure}[]
\begin{minipage}{\textwidth}
\begin{algorithm}[H]
\caption{Class Generation}\label{alg:class-generation}
\begin{algorithmic}[1]
\State Initialize the set of unconnected classes: $U = \{C_{1}, C_{2}, ..., C_{m}\}$
\State Initialize the set of linked classes: $L = \emptyset$
\State $C_{1} \gets \Call{U.pop}{\null}$, $\Call{L.add}{C_{1}}$
\State $C_{1}$ \texttt{rdfs:subClassOf} \texttt{root}
\While{\texttt{max\_depth} is not satisfied} \Comment{Fulfilling class hierarchy depth}
    \State $C_{i} \gets \Call{U.pop}{\null}$, $\Call{L.add}{C_{i}}$
    \State $C_{i}$ \texttt{rdfs:subClassOf} $C_{i-1}$
\EndWhile
\While{$U \neq  \emptyset$} \Comment{Building class hierarchy}
    \State $C_{i} \gets \Call{U.pop}{\null}$, $\Call{L.add}{C_{i}}$
    \If{$\text{random}(0, 1) \leq \alpha$} \Comment{Adding stochasticity}
        \State Place it randomly
    \Else
        \State Place it s.t. \texttt{avg\_depth} and \texttt{inheritance\_ratio} are satisfied
    \EndIf
\EndWhile
\While{\texttt{current\_disj} $<$ \texttt{disj\_ratio}} \Comment{Adds \texttt{owl:disjointWith}}
    \State Pick classes A and B s.t. B is neither a parent nor a child of A
    \State Make A and B disjoint and extend disjointness to their respective children
\EndWhile
\end{algorithmic}
\end{algorithm}

\begin{algorithm}[H]
\caption{Relation Generation}\label{alg:relation-generation}
\begin{algorithmic}[1]
\State Initialize the set of unqualified relations: $U = \{R_{1}, R_{2}, ..., R_{n}\}$
\State Initialize the set of qualified relations: $Q = \emptyset $
\State Load compatible patterns
\While{$U \neq \emptyset$ and attribute proportions not satisfied} \Comment{Adding attributes}
    \State $R_{i} \gets \Call{U.pop}{\null}$, $\Call{Q.add}{R_{i}}$
    \State $\Call{qualify}{R_{i}}$ based on priority order
\EndWhile
\State $\Call{inverse\_pairing}{Q}$ \Comment{Adding \texttt{owl:inverseOf}}
\State $\Call{relation\_profiling}{Q}$ \Comment{Adding \texttt{rdfs:domain/range}}
\State $\Call{subproperty\_pairing}{Q}$ \Comment{Adding \texttt{rdfs:subPropertyOf}}
\end{algorithmic}
\end{algorithm}

\begin{algorithm}[H]
\caption{Knowledge Graph Generation}\label{alg:kg-generation}
\begin{algorithmic}[1]
\State Initialize the set of unobserved entities: $U = \{E_{1}, E_{2}, ..., E_{p}\}$
\State Initialize the set of observed entities: $O = \emptyset$
\State Initialize the knowledge graph: $\mathcal{KG} = \emptyset$
\State Load the underpinning schema $\mathcal{S}$
\State $\Call{assign\_class}{U}$ \Comment{Specific class attribution based on \texttt{prop\_untyped}}
\If{\texttt{multityping}}
    \State $\Call{complete\_typing}{U}$ \Comment{Adding specific classes based on \texttt{avg\_multityping}}
\EndIf
\While{$U \neq \emptyset$ and \texttt{num\_triples} not satisfied}
    \State $\mathcal{KG} \gets \Call{generate\_triples}{U, O, \mathcal{S}}$
\EndWhile
\State $\Call{checking\_procedures}{\mathcal{KG}}$ \Comment{Removing foreseeable inconsistencies}
\State $\Call{reasoning}{\mathcal{KG}}$ \Comment{HermiT reasoner}
\end{algorithmic}
\end{algorithm}

\end{minipage}
\end{figure}

\section{PyGraft in Action}\label{pygraft-use-case}
\subsection{Efficiency and Scalability Details}
In this section, the efficiency and scalability of PyGraft are benchmarked across several schema and graph configurations. Each schema specification reported in Table~\ref{tab:schemas-specs} is paired with each graph specification from Table~\ref{tab:kg-specs}. This leads to 27 distinct combinations. 

In particular, schemas from $\mathcal{S}1$ to $\mathcal{S}3$ are small-sized, schemas from $\mathcal{S}4$ to $\mathcal{S}6$ are medium-sized, and schemas from $\mathcal{S}7$ to $\mathcal{S}9$ are of larger sizes (Table~\ref{tab:schemas-specs}). For each schema of a given size, the degree of constraints vary as they contain different levels of \texttt{OWL} and \texttt{RDFS} logical constructs. For example, $\mathcal{S}1$ has less constraints than $\mathcal{S}2$, which itself has less constraints than $\mathcal{S}3$.
Graph specifications $\mathcal{G}_1$, $\mathcal{G}_2$, and $\mathcal{G}_3$ correspond to small-sized, medium-sized and large-sized graphs, respectively (Table~\ref{tab:kg-specs}).

\begin{figure}[h]
  \centering
  \includegraphics[scale=0.35]{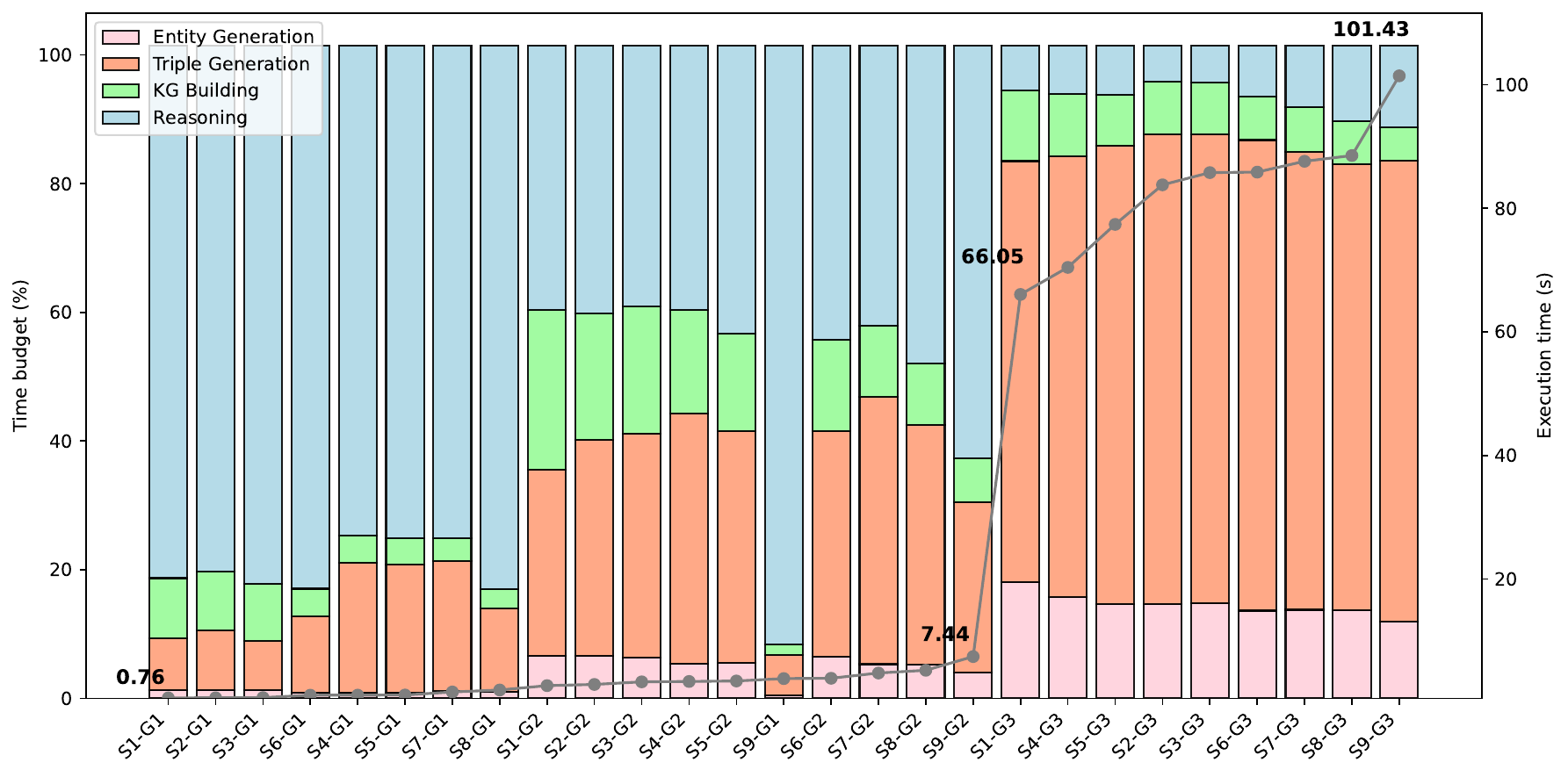}
  \caption{Execution time breakdown for each configuration.}
  \label{fig:barplots}
\end{figure}

For these 27 unique configurations, execution times w.r.t. several dimensions are computed and shown in Fig.~\ref{fig:barplots}. Execution times related to the schema generation are omitted as they are negligible. Experiments were conducted on a machine with 2 CPUs Intel Xeon E5-2650 v4, 12 cores/CPU, and 128GB RAM.

It is worth mentioning that the 27 generated KGs were flagged as consistent at the first attempt. The breakdown of time executions differs according to the schema and graph sizes (see Fig.~\ref{fig:barplots}). For small graphs, the final consistency checking is the most time-consuming part. For large graphs, triple generation time dominates the rest. 
As graph sizes increase, all execution times grow but we observe that PyGraft is able to generate consistent KGs quickly, even for large KGs: with our experimental configuration, the total execution time for KGs with 10K entities and 100K triples is roughly 1.5 minutes. In addition, PyGraft scalability was assessed by asking a KG of 100K entities and 1M triples. On the same machine, it took 47 minutes to generate such a KG, which was again consistent at the first attempt.

\begin{table}[htbp]
\caption{Generated schemas. Column headers from left to right: number of classes, class hierarchy depth, average class depth, proportion of class disjointness (\textit{cd}), number of relations, average depth of relation domains and ranges (\textit{rs}), and proportions of reflexive (\textit{rf}), irreflexive (\textit{irr}), asymmetric (\textit{asy}), symmetric (\textit{sy}), transitive (\textit{tra}), and inverse (\textit{inv}) relations.}
\label{tab:schemas-specs}
\centering
\setlength{\tabcolsep}{0.25cm}
\resizebox{1.0\textwidth}{!}{
\begin{tabular}{cccccccccccccc}
\toprule 
& $|\mathcal{C}|$ & $\operatorname{MAX}(\mathcal{D})$ & $\operatorname{AVG}(\mathcal{D})$ & \textit{cd} & & $|\mathcal{R}|$ & \textit{rs} & \textit{ref} & \textit{irr} & \textit{asy} & \textit{sym} & \textit{tra} & \textit{inv}\\
\midrule
$\mathcal{S}1$ & $25$ & $3$ & $1.5$ & $0.1$ & & $25$ & $1.5$ & $0.1$ & $0.1$ & $0.1$ & $0.1$ & $0.1$ & $0.1$ \\
\midrule
$\mathcal{S}2$ & $25$ & $3$ & $1.5$ & $0.2$ & & $25$ & $1.5$ & $0.2$ & $0.2$ & $0.2$ & $0.2$ & $0.2$ & $0.2$ \\
\midrule
$\mathcal{S}3$ & $25$ & $3$ & $1.5$ & $0.3$ & & $25$ & $1.5$ & $0.3$ & $0.3$ & $0.3$ & $0.3$ & $0.3$ & $0.3$ \\
\midrule
$\mathcal{S}4$ & $100$ & $4$ & $2.5$ & $0.1$ & & $100$ & $2.5$ & $0.1$ & $0.1$ & $0.1$ & $0.1$ & $0.1$ & $0.1$ \\
\midrule
$\mathcal{S}5$ & $100$ & $4$ & $2.5$ & $0.2$ & & $100$ & $2.5$ & $0.2$ & $0.2$ & $0.2$ & $0.2$ & $0.2$ & $0.2$ \\
\midrule
$\mathcal{S}6$ & $100$ & $4$ & $2.5$ & $0.3$ & & $100$ & $2.5$ & $0.3$ & $0.3$ & $0.3$ & $0.3$ & $0.3$ & $0.3$ \\
\midrule
$\mathcal{S}7$ & $250$ & $5$ & $3.0$ & $0.1$ & & $250$ & $3.0$ & $0.1$ & $0.1$ & $0.1$ & $0.1$ & $0.1$ & $0.1$ \\
\midrule
$\mathcal{S}8$ & $250$ & $5$ & $3.0$ & $0.2$ & & $250$ & $3.0$ & $0.2$ & $0.2$ & $0.2$ & $0.2$ & $0.2$ & $0.2$ \\
\midrule
$\mathcal{S}9$ & $250$ & $5$ & $3.0$ & $0.3$ & & $250$ & $3.0$ & $0.3$ & $0.3$ & $0.3$ & $0.3$ & $0.3$ & $0.3$ \\
\bottomrule
\end{tabular}}
\end{table}

\begin{table}[htbp]
\caption{Different graph specifications. Column headers from left to right: number of entities, number of triples, proportion of untyped entities, average depth of the most specific specific class, average number of most-specific classes per multi-typed entity.}
\label{tab:kg-specs}
\centering
\setlength{\tabcolsep}{0.25cm}
\resizebox{0.5\textwidth}{!}{
\begin{tabular}{cccccc}
\toprule 
& $|\mathcal{E}|$ & $|\mathcal{T}|$ & \textit{unt} & \textit{asc} & \textit{mul}\\
\midrule
$\mathcal{G}_1$ & $100$ & $1,000$ & $0.3$ & $2.0$ & $2.0$ \\
\midrule
$\mathcal{G}_2$ & $1,000$ & $10,000$ & $0.3$ & $2.0$ & $2.0$ \\
\midrule
$\mathcal{G}_3$ & $10,000$ & $100,000$ & $0.3$ & $2.0$ & $2.0$ \\
\bottomrule
\end{tabular}}
\end{table}

\subsection{Usage Illustration}
In this section, we briefly provide some usage examples to demonstrate how easy it is to use PyGraft. Desired characteristics of the output schema and/or KG can be specified with the path to a \texttt{yaml} or \texttt{json} configuration file. In the example presented in Listing~\ref{list:usage}, after importing PyGraft, we first generate a \texttt{yaml} configuration file in the current working directory. For the sake of simplicity, the generated template is left untouched, \textit{i.e.,} we keep the default parameter values. More advanced usage is provided in the official documentation: \url{https://pygraft.readthedocs.io/en/latest/}.
Then, we generate both a schema and KG in a single pipeline. The generated resources are subsequently stored in the current working directory.

\begin{lstlisting}[style=pythonstyle, caption={Schema and KG generation with PyGraft.}, label={list:usage}]
import pygraft

pygraft.create_yaml_template()
pygraft.generate("template.yml")
\end{lstlisting}

By default, the generated graph is stored as an \texttt{rdf/xml} file. A snippet of a KG generated with PyGraft is provided in Listing~\ref{list:snippet}.

\begin{lstlisting}[caption={Excerpt from a generated graph.}, label={list:snippet}, captionpos=b, style=customxml]
<?xml version="1.0" encoding="utf-8"?>
<rdf:RDF
   xmlns:ns1="http://purl.org/dc/terms/"
   xmlns:owl="http://www.w3.org/2002/07/owl#"
   xmlns:rdf="http://www.w3.org/1999/02/22-rdf-syntax-ns#"
   xmlns:rdfs="http://www.w3.org/2000/01/rdf-schema#"
   xmlns:sc="http://pygraf.t/"
>

  <rdf:Description rdf:about="http://pygraf.t/C30">
    <rdf:type rdf:resource="http://www.w3.org/2002/07/owl#Class"/>
    <rdfs:subClassOf rdf:resource="http://www.w3.org/2002/07/owl#Thing"/>
    <owl:disjointWith rdf:resource="http://pygraf.t/C16"/>
  </rdf:Description>
  <rdf:Description rdf:about="http://pygraf.t/E324">
    <rdf:type rdf:resource="http://pygraf.t/C4"/>
    <rdf:type rdf:resource="http://pygraf.t/C17"/>
    <schema:R15 rdf:resource="http://pygraf.t/E356"/>
    <schema:R34 rdf:resource="http://pygraf.t/E44"/>
  </rdf:Description>
  <rdf:Description rdf:about="http://pygraf.t/R34">
    <rdf:type rdf:resource="http://www.w3.org/2002/07/owl#ObjectProperty"/>
    <rdf:type rdf:resource="http://www.w3.org/2002/07/owl#TransitiveProperty"/>
    <rdfs:domain rdf:resource="http://pygraf.t/C17"/>
    <rdfs:range rdf:resource="http://pygraf.t/C17"/>
    <owl:inverseOf rdf:resource="http://pygraf.t/R15"/>
  </rdf:Description>
\end{lstlisting}

\section{Discussion}\label{discussion}
\subsection{Potential Uses}
As mentioned, PyGraft can be used for generating several schemas and KGs on the fly, thereby facilitating novel approaches and model benchmarking on a wider range of datasets with diversified characteristics. In practice, this can be easily done with inspecting the configuration file we provide as template, and tweaking parameters as one sees fit.

Additionally, PyGraft can be used for generating anonymous data. Let us assume one is working in a data-sensitive field such as medicine or education, characterized by the paucity of readily available data. Thanks to PyGraft, it becomes much more accessible to experiment with sensitiveless testbeds, provided that the user has a rough idea of the scale and characteristics of the resources they want to mimic.

It is worth noting that a vibrant research community aims at incorporating schema-based information into the learning process with KGs~\cite{damato2021,hubert2023-maschine,jain2021}. Using information derived from the schema in addition to the facts contained in the KG aims at building more semantic-aware approaches~\cite{hubert2023-loss}, that are expected to result in more coherent predictions~\cite{hubert2023-sem} and better predictive accuracy~\cite{damato2021,hubert2023-loss}. However, it is frequently reported that most KGs do not come with publicly available schemas that would facilitate the development of such schema-aware approaches. For this reason, PyGraft may also be helpful to generate both synthetic schemas with dependent KGs to foster the development of schema-driven, neuro-symbolic approaches, and to run ablation studies on specific schema constructs in isolation.

\subsection{Limitations, Sustainability, Maintenance and Future Work}
In this section, we discuss the current limitations of PyGraft and elaborate on our sustainability and maintenance plan, as well as future work opportunities.

First, it should be noted that PyGraft relies on \texttt{rdflib} for serializing triples. As part of supplementary experiments, we tried to push PyGraft capabilities to its limit by generating very large KGs with >10M entities and triples. In such cases, the serialization failed. In future work, we will develop independent serialization procedures so that bigger graphs can be generated.

Secondly, recall that our implemented checking procedures limit the likelihood of any inconsistency \emph{before} the DL reasoner is ultimately applied on the generated KG. In case of inconsistencies that would remain undetected by the checking procedures, our actual use of the HermiT reasoner is able to detect such inconsistencies, but not to provide any information on the triples that should be removed so that the KG becomes consistent. In future work, we will implement such a functionality, so that the consistency of the generated KGs can be ensured in a single loop without requiring any input from the user's side.

More generally, we aim at maintaining the PyGraft library, a cornerstone in our commitment to robust software engineering practices and open-source community engagement. Our approach is to release new versions in response to emerging user requirements, ensuring that PyGraft not only meets but anticipates the needs of its users. For instance, a common request is the ability to generate literals -- which we seek to achieve in the near future. Furthermore, we recently welcomed the suggestion of offering a hub that would gather different KG profiles generated by the community.
To this aim, we welcome all contributions on Github (\textit{e.g.,}, issues, forks, or pull requests) to make PyGraft useful to researchers of a large array of communities.
We are dedicated to incorporating methodologies for continuous integration and deployment, fostering a test-driven development environment that guarantees reliability and efficiency. We also want PyGraft to fit within the Semantic Web community and be recognized as a useful tool for researchers and engineers working on KG-based applications.

In future work, we will showcase the impact of PyGraft in real-world use cases. Some relevant scenarios include recommender systems and ontology repairment. In particular, for the latter use-case, we envision the creation of voluntarily imperfect schemas, containing (known) contradictions to resolve. Link prediction also appears as an appropriate task to evaluate PyGraft's usefulness.

\section{Conclusion}\label{conclusion}
In this work, we presented PyGraft, a Python tool for generating synthetic schemas and KGs from user requirements. Several \texttt{OWL} and \texttt{RDFS} constructs are integrated to output realistic KGs that comply with Semantic Web standards. PyGraft allows researchers and practitioners to generate schemas and KGs on the fly, provided minimal knowledge about the desired specifications. Hence, PyGraft can prove useful for various applications. As it allows for generating schemas and KGs of controlled characteristics, it can serve as a well-suited tool for benchmarking novel approaches or models. Being domain-agnostic, PyGraft can be used to generate synthetic schemas and KGs resembling real ones in data-sensitive fields where access or publication of public data is scarce. Due to the richly described generated KGs in terms of \texttt{OWL} and \texttt{RDFS} constructs, PyGraft may also facilitate the development of schema-driven, neuro-symbolic models.

\clearpage
%
%
%
\bibliographystyle{splncs04}
\bibliography{bibliography}

\begin{thebibliography}{10}
\providecommand{\url}[1]{\texttt{#1}}
\providecommand{\urlprefix}{URL }
\providecommand{\doi}[1]{https://doi.org/#1}

\bibitem{barabasi2002}
Albert, R., Barab\'asi, A.L.: Statistical mechanics of complex networks. Rev. Mod. Phys.  \textbf{74},  47--97 (Jan 2002). \doi{10.1103/RevModPhys.74.47}

\bibitem{angles2014}
Angles, R., Boncz, P.A., Larriba{-}Pey, J.L., Fundulaki, I., Neumann, T., Erling, O., Neubauer, P., Mart{\'{\i}}nez{-}Bazan, N., Kotsev, V., Toma, I.: The linked data benchmark council: a graph and {RDF} industry benchmarking effort. {SIGMOD} Rec.  \textbf{43}(1),  27--31 (2014). \doi{10.1145/2627692.2627697}

\bibitem{bagan2017}
Bagan, G., Bonifati, A., Ciucanu, R., Fletcher, G.H.L., Lemay, A., Advokaat, N.: gmark: Schema-driven generation of graphs and queries. {IEEE} Trans. Knowl. Data Eng.  \textbf{29}(4),  856--869 (2017). \doi{10.1109/TKDE.2016.2633993}

\bibitem{bonatti}
Bonatti, P.A., Decker, S., Polleres, A., Presutti, V.: Knowledge graphs: New directions for knowledge representation on the semantic web (dagstuhl seminar 18371). Dagstuhl Reports  \textbf{8}(9),  29--111 (2018). \doi{10.4230/DagRep.8.9.29}

\bibitem{rmat2004}
Chakrabarti, D., Zhan, Y., Faloutsos, C.: {R-MAT:} {A} recursive model for graph mining. In: Proceedings of the Fourth {SIAM} International Conference on Data Mining, Lake Buena Vista, Florida, USA, April 22-24, 2004. pp. 442--446. {SIAM} (2004). \doi{10.1137/1.9781611972740.43}

\bibitem{damato2021}
d'Amato, C., Quatraro, N.F., Fanizzi, N.: Injecting background knowledge into embedding models for predictive tasks on knowledge graphs. In: The Semantic Web - 18th International Conference, {ESWC} 2021, Virtual Event, June 6-10, 2021, Proceedings. Lecture Notes in Computer Science, vol. 12731, pp. 441--457. Springer (2021). \doi{10.1007/978-3-030-77385-4\_26}

\bibitem{molgan2018}
De~Cao, N., Kipf, T.: {MolGAN: An implicit generative model for small molecular graphs}. ICML 2018 workshop on Theoretical Foundations and Applications of Deep Generative Models  (2018)

\bibitem{dettmers2018}
Dettmers, T., Minervini, P., Stenetorp, P., Riedel, S.: Convolutional 2d knowledge graph embeddings. In: Proceedings of the Thirty-Second {AAAI} Conference on Artificial Intelligence, (AAAI-18), the 30th innovative Applications of Artificial Intelligence (IAAI-18), and the 8th {AAAI} Symposium on Educational Advances in Artificial Intelligence (EAAI-18), New Orleans, Louisiana, USA, February 2-7, 2018. pp. 1811--1818. {AAAI} Press (2018)

\bibitem{ehrlinger}
Ehrlinger, L., W{\"{o}}{\ss}, W.: Towards a definition of knowledge graphs. In: Joint Proceedings of the Posters and Demos Track of the 12th International Conference on Semantic Systems - SEMANTiCS2016 and the 1st International Workshop on Semantic Change {\&} Evolving Semantics (SuCCESS'16) co-located with the 12th International Conference on Semantic Systems (SEMANTiCS 2016), Leipzig, Germany, September 12-15, 2016. {CEUR} Workshop Proceedings, vol.~1695. CEUR-WS.org (2016)

\bibitem{erdos1959}
ERDdS, P., R\&wi, A.: On random graphs i. Publ. math. debrecen  \textbf{6}(290-297), ~18 (1959)

\bibitem{feng2021}
Feng, Z., Mayer, W., He, K., Kwashie, S., Stumptner, M., Grossmann, G., Peng, R., Huang, W.: A schema-driven synthetic knowledge graph generation approach with extended graph differential dependencies (gdd\({}^{\mbox{x}}\)s). {IEEE} Access  \textbf{9},  5609--5639 (2021). \doi{10.1109/ACCESS.2020.3048186}

\bibitem{hermit}
Glimm, B., Horrocks, I., Motik, B., Stoilos, G., Wang, Z.: Hermit: An {OWL} 2 reasoner. J. Autom. Reason.  \textbf{53}(3),  245--269 (2014). \doi{10.1007/s10817-014-9305-1}

\bibitem{graphgen2020}
Goyal, N., Jain, H.V., Ranu, S.: Graphgen: {A} scalable approach to domain-agnostic labeled graph generation. In: {WWW} '20: The Web Conference 2020, Taipei, Taiwan, April 20-24, 2020. pp. 1253--1263. {ACM} / {IW3C2} (2020). \doi{10.1145/3366423.3380201}

\bibitem{gruber1995}
Gruber, T.R.: Toward principles for the design of ontologies used for knowledge sharing? Int. J. Hum. Comput. Stud.  \textbf{43}(5-6),  907--928 (1995). \doi{10.1006/ijhc.1995.1081}

\bibitem{guo2005}
Guo, Y., Pan, Z., Heflin, J.: {LUBM:} {A} benchmark for {OWL} knowledge base systems. J. Web Semant.  \textbf{3}(2-3),  158--182 (2005). \doi{10.1016/j.websem.2005.06.005}

\bibitem{kgbook}
Hogan, A., Blomqvist, E., Cochez, M., d'Amato, C., de~Melo, G., Gutierrez, C., Kirrane, S., Gayo, J.E.L., Navigli, R., Neumaier, S., Ngomo, A.N., Polleres, A., Rashid, S.M., Rula, A., Schmelzeisen, L., Sequeda, J., Staab, S., Zimmermann, A.: Knowledge Graphs. Synthesis Lectures on Data, Semantics, and Knowledge, Morgan {\&} Claypool Publishers (2021). \doi{10.2200/S01125ED1V01Y202109DSK022}

\bibitem{hubert2023-loss}
Hubert, N., Monnin, P., Brun, A., Monticolo, D.: Enhancing knowledge graph embedding models with semantic-driven loss functions. CoRR  \textbf{abs/2303.00286} (2023). \doi{10.48550/arXiv.2303.00286}

\bibitem{hubert2023-sem}
Hubert, N., Monnin, P., Brun, A., Monticolo, D.: Sem@k: Is my knowledge graph embedding model semantic-aware? CoRR  \textbf{abs/2301.05601} (2023). \doi{10.48550/arXiv.2301.05601}

\bibitem{hubert2023-maschine}
Hubert, N., Paulheim, H., Monnin, P., Brun, A., Monticolo, D.: Schema first! learn versatile knowledge graph embeddings by capturing semantics with maschine. CoRR  \textbf{abs/2306.03659} (2023). \doi{10.48550/arXiv.2306.03659}

\bibitem{jain2021}
Jain, N., Tran, T., Gad{-}Elrab, M.H., Stepanova, D.: Improving knowledge graph embeddings with ontological reasoning. In: The Semantic Web - {ISWC} 2021 - 20th International Semantic Web Conference, {ISWC} 2021, Virtual Event, October 24-28, 2021, Proceedings. Lecture Notes in Computer Science, vol. 12922, pp. 410--426. Springer (2021). \doi{10.1007/978-3-030-88361-4\_24}

\bibitem{jin2023}
Jin, L., Yao, Z., Chen, M., Chen, H., Zhang, W.: A comprehensive study on knowledge graph embedding over relational patterns based on rule learning (2023)

\bibitem{liu2023}
Liu, S., Grau, B.C., Horrocks, I., Kostylev, E.V.: Revisiting inferential benchmarks for knowledge graph completion. CoRR  \textbf{abs/2306.04814} (2023). \doi{10.48550/arXiv.2306.04814}

\bibitem{melo2017}
Melo, A., Paulheim, H.: Synthesizing knowledge graphs for link and type prediction benchmarking. In: The Semantic Web - 14th International Conference, {ESWC} 2017, Portoro{\v{z}}, Slovenia, May 28 - June 1, 2017, Proceedings, Part {I}. Lecture Notes in Computer Science, vol. 10249, pp. 136--151 (2017). \doi{10.1007/978-3-319-58068-5\_9}

\bibitem{palowitch2022}
Palowitch, J., Tsitsulin, A., Mayer, B., Perozzi, B.: Graphworld: Fake graphs bring real insights for gnns. In: {KDD} '22: The 28th {ACM} {SIGKDD} Conference on Knowledge Discovery and Data Mining, Washington, DC, USA, August 14 - 18, 2022. pp. 3691--3701. {ACM} (2022). \doi{10.1145/3534678.3539203}

\bibitem{trilliong2017}
Park, H., Kim, M.: Trilliong: {A} trillion-scale synthetic graph generator using a recursive vector model. In: Proceedings of the 2017 {ACM} International Conference on Management of Data, {SIGMOD} Conference 2017, Chicago, IL, USA, May 14-19, 2017. pp. 913--928. {ACM} (2017). \doi{10.1145/3035918.3064014}

\bibitem{portisch2022}
Portisch, J., Paulheim, H.: The {DLCC} node classification benchmark for analyzing knowledge graph embeddings. In: The Semantic Web - {ISWC} 2022 - 21st International Semantic Web Conference, Virtual Event, October 23-27, 2022, Proceedings. Lecture Notes in Computer Science, vol. 13489, pp. 592--609. Springer (2022). \doi{10.1007/978-3-031-19433-7\_34}

\bibitem{rossi2021}
Rossi, A., Firmani, D., Merialdo, P., et~al.: Knowledge graph embeddings or bias graph embeddings? a study of bias in link prediction models. In: CEUR WORKSHOP PROCEEDINGS. vol.~3034. CEUR-WS (2021)

\bibitem{rossi2020relations}
Rossi, A., Matinata, A.: Knowledge graph embeddings: Are relation-learning models learning relations? In: Proceedings of the Workshops of the {EDBT/ICDT} 2020 Joint Conference, Copenhagen, Denmark, March 30, 2020. {CEUR} Workshop Proceedings, vol.~2578. CEUR-WS.org (2020)

\bibitem{nevae2020}
Samanta, B., De, A., Jana, G., G{\'{o}}mez, V., Chattaraj, P.K., Ganguly, N., Gomez{-}Rodriguez, M.: {NEVAE:} {A} deep generative model for molecular graphs. J. Mach. Learn. Res.  \textbf{21},  114:1--114:33 (2020)

\bibitem{graphvae2018}
Simonovsky, M., Komodakis, N.: Graphvae: Towards generation of small graphs using variational autoencoders. In: Artificial Neural Networks and Machine Learning - {ICANN} 2018 - 27th International Conference on Artificial Neural Networks, Rhodes, Greece, October 4-7, 2018, Proceedings, Part {I}. Lecture Notes in Computer Science, vol. 11139, pp. 412--422. Springer (2018). \doi{10.1007/978-3-030-01418-6\_41}

\bibitem{graphgan2018}
Wang, H., Wang, J., Wang, J., Zhao, M., Zhang, W., Zhang, F., Xie, X., Guo, M.: Graphgan: Graph representation learning with generative adversarial nets. In: Proceedings of the Thirty-Second {AAAI} Conference on Artificial Intelligence, (AAAI-18), the 30th innovative Applications of Artificial Intelligence (IAAI-18), and the 8th {AAAI} Symposium on Educational Advances in Artificial Intelligence (EAAI-18), New Orleans, Louisiana, USA, February 2-7, 2018. pp. 2508--2515. {AAAI} Press (2018)

\bibitem{graphrnn2018}
You, J., Ying, R., Ren, X., Hamilton, W.L., Leskovec, J.: Graphrnn: Generating realistic graphs with deep auto-regressive models. In: Proceedings of the 35th International Conference on Machine Learning, {ICML} 2018, Stockholmsm{\"{a}}ssan, Stockholm, Sweden, July 10-15, 2018. Proceedings of Machine Learning Research, vol.~80, pp. 5694--5703. {PMLR} (2018)

\end{thebibliography}
\end{document}